\documentclass[letterpaper]{article} 
\usepackage[]{aaai25}  
\usepackage{times}  
\usepackage{helvet}  
\usepackage{courier}  
\usepackage[hyphens]{url}  
\usepackage{graphicx} 
\usepackage{natbib}  
\usepackage{caption} 
\usepackage{algorithm}
\usepackage{multirow}
\usepackage{amsmath}
\usepackage{algpseudocode}

\usepackage{newfloat}
\usepackage{listings}
\DeclareCaptionStyle{ruled}{labelfont=normalfont,labelsep=colon,strut=off} 
\floatstyle{ruled}
\newfloat{listing}{tb}{lst}{}
\floatname{listing}{Listing}
\title{VERA: Validation and Enhancement for Retrieval Augmented systems}

\author{
Nitin Aravind Birur, Tanay Baswa, Divyanshu Kumar, Jatan Loya, Sahil Agarwal, Prashanth Harshangi 
}
\affiliations{
    Enkrypt AI, Boston, MA, USA \\

    \{nitin, tanay, divyanshu, jatan,  sahil, prashanth\}@enkryptai.com
}

\usepackage{bibentry}

\begin{document}

\maketitle

\begin{abstract}
Large language models (LLMs) exhibit remarkable capabilities but often produce inaccurate responses, as they rely solely on their embedded knowledge. Retrieval-Augmented Generation (RAG) enhances LLMs by incorporating an external information retrieval system, supplying additional context along with the query to mitigate inaccuracies for a particular context. However, accuracy issues still remain, as the model may rely on irrelevant documents or extrapolate incorrectly from its training knowledge. To assess and improve the performance of both the retrieval system and the LLM in a RAG framework, we propose \textbf{VERA} (\textbf{V}alidation and \textbf{E}nhancement for \textbf{R}etrieval \textbf{A}ugmented systems), a system designed to: 1) Evaluate and enhance the retrieved context before response generation, and 2) Evaluate and refine the LLM-generated response to ensure precision and minimize errors. VERA employs an evaluator-cum-enhancer LLM that first checks if external retrieval is necessary, evaluates the relevance and redundancy of the retrieved context, and refines it to eliminate non-essential information. Post-response generation, VERA splits the response into atomic statements, assesses their relevance to the query, and ensures adherence to the context. Our experiments demonstrate VERA’s remarkable efficacy not only in improving the performance of smaller open-source models, but also larger state-of-the art models. These enhancements underscore VERA’s potential to produce accurate and relevant responses, advancing the state-of-the-art in retrieval-augmented language modeling. VERA's robust methodology, combining multiple evaluation and refinement steps, effectively mitigates hallucinations and improves retrieval and response processes, making it a valuable tool for applications demanding high accuracy and reliability in information generation.
\end{abstract}

\section{Introduction}
\begin{figure}[ht]
    \centering
    \includegraphics[width=\columnwidth]{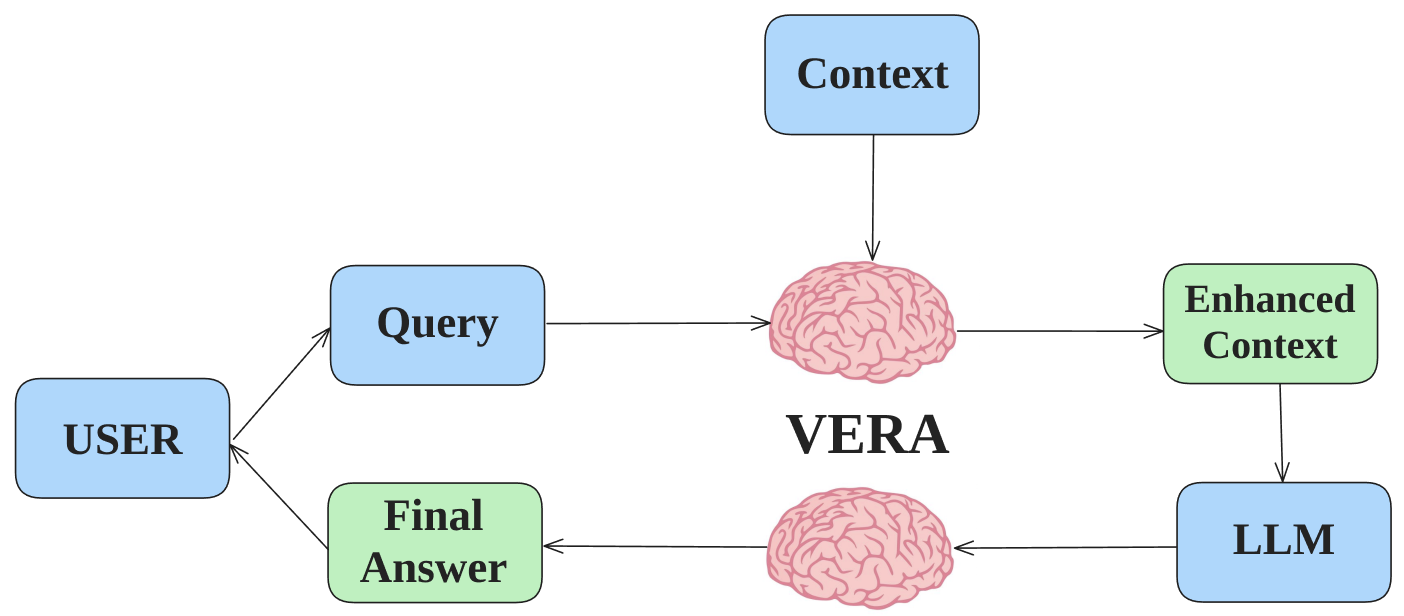}
    \caption{An overview of VERA}
\end{figure}
Retrieval-Augmented Generation (RAG) \citep{NEURIPS2020_6b493230} techniques enhance the inputs to Large Language Models (LLMs) by incorporating relevant retrieved passages, thus reducing factual errors in knowledge-intensive tasks. These passages are retrieved using methods such as vector similarity search. However, previous research has demonstrated that retrieval-augmented models may generate text that includes additional information beyond the retrieved documents \citep{dziri2022origin}, disregards the documents altogether \citep{krishna2021hurdles}, or even contradicts the documents \citep{longpre2021entity}. The quality of the LLM’s response can also be compromised by erroneous or irrelevant retrievals \citep{khandelwal2019generalization}. In reality, retrievals are not always necessary and are primarily needed for knowledge-intensive tasks. Therefore, there is a critical need to enhance both the quality of retrievals and the quality of responses.

To quantify and evaluate the quality of retrievals and responses, we employ the following metrics:
\begin{itemize}
    \item \textbf{Response Adherence:} This metric measures the extent to which the LLM’s response is grounded in the provided context.
    \item \textbf{Response Relevance:} This metric evaluates the amount of information in the LLM’s response that is relevant to and helps in answering the given query.
    \item \textbf{Context Relevance:} This metric assesses the amount of information in the retrieved context that is pertinent to and aids in answering the given query.
\end{itemize}

These metrics allow for a comprehensive evaluation of both the retrieval process and the subsequent response generation, ensuring improvements in the overall performance of RAG systems.

VERA enhances the Context Relevance of retrieved sources prior to their input into the LLM and subsequently improves the Response Adherence and Relevance after the LLM generates its response. To achieve this, VERA employs an evaluator-cum-enhancer LLM that assesses the content, utilizing reasoning to determine optimal edits, which are then executed while preserving the original structure and style of both the context and the response as much as possible.

In our effort to enhance the performance of RAG systems with any arbitrary retrieval system and LLM, we contribute the following advancements:

\begin{enumerate}
    \item \textbf{Robust and Fine-Grained Evaluation Technique:} We introduce a comprehensive evaluation method to assess any given retrieval system and LLM using the previously mentioned metrics.
    \item \textbf{System for Context and Response Enhancement:} We propose a system that leverages the fine-grained evaluation results to analyze and perform appropriate edits to the context (before response generation) and the response. The ultimate goal is to produce error-free, relevant responses using a RAG system.
\end{enumerate}
Moreover, our method is designed to be easily reproducible, allowing seamless integration into any existing RAG system.

\section{Related Works}

\subsection{RARR}

The RARR \citep{gao2023rarrresearchingrevisinglanguage} framework retroactively enables large language models (LLMs) to attribute external evidence through a process termed Editing for Attribution. Given a model-generated text, RARR conducts a research stage to locate evidence supporting the text's statements. Subsequently, in the revision stage, the framework utilizes this gathered evidence to amend any facts in the original text that lack support, while striving to preserve the initial content as much as possible. RARR primarily aims to correct and attribute model-generated texts within open domain scenarios that lack supporting context in the input prompt. Although this approach can be applied to closed-domain retrieval-augmented generation (RAG) pipelines, it does not enhance the relevance of the context or answers.

\subsection{SELF-RAG}

The Self-RAG framework, as introduced by \citep{asai2023self}, represents a pioneering approach in natural language generation (NLG) by integrating self-reflection mechanisms into the training and generation process of a language model (LM). This end-to-end trained LM generates output in segmented form, guided by specialized reflection tokens designed to enhance its performance. Key among these tokens is the Retrieve token, which determines whether the model should retrieve multiple documents in parallel to inform its generation process. If retrieval is activated (Retrieve == yes), the model evaluates the relevance of retrieved documents using the IsRel token. This token categorizes relevance as either "relevant" or "irrelevant," thereby assisting the model in selecting pertinent information. Subsequently, the IsSup token assesses the degree to which the generated output is supported by the retrieved documents, while the IsUse token judges the usefulness of the generated text on a predefined scale. By iteratively applying these tokens, Self-RAG aims to improve the quality, relevance, and utility of its generated outputs through self-critique and refinement. However, SELF-RAG is not very flexible or versatile as training a language model is both resource-intensive and time-consuming.

\subsection{CRAG}
The Corrective RAG (CRAG) paper \citep{yan2024corrective} introduces a method to enhance the accuracy of language models by reintegrating information from retrieved documents. It employs an evaluator to assess the quality of the documents obtained for a query and then determines whether to use, ignore, or request additional data from these documents. CRAG also utilizes web searches to expand its information beyond static databases, ensuring access to a broader, up-to-date range of information. Additionally, it employs a unique strategy to decompose and reconstruct retrieved documents, emphasizing the extraction of the most relevant information while eliminating distractions. Although CRAG improves the quality of retrieval, it does not address inaccuracies and irrelevancies in the final response. While CRAG’s ability to access the web for external information may be useful for general-purpose question answering, most critical applications of RAG systems aim to limit the LLM's scope of knowledge to the provided documents (e.g., customer service bots).

\subsection{FACTScore}
FACTSCORE \citep{min2023factscore} introduces a method to evaluate the factual accuracy of language models by decomposing their outputs into atomic facts and verifying each one against a specified knowledge source. It also presents a model that approximates FACTSCORE with an error rate of less than 2\%, enabling the evaluation of a large set of new LMs without requiring manual human effort. VERA employs a similar technique to assess the context adherence of responses. However, FACTSCORE is purely an evaluation technique for testing adherence quality and does not address the quality enhancement of context retrieval or the responses.

\section{Methodology}
\begin{figure}[ht]
    \centering
    \includegraphics[width=\columnwidth]{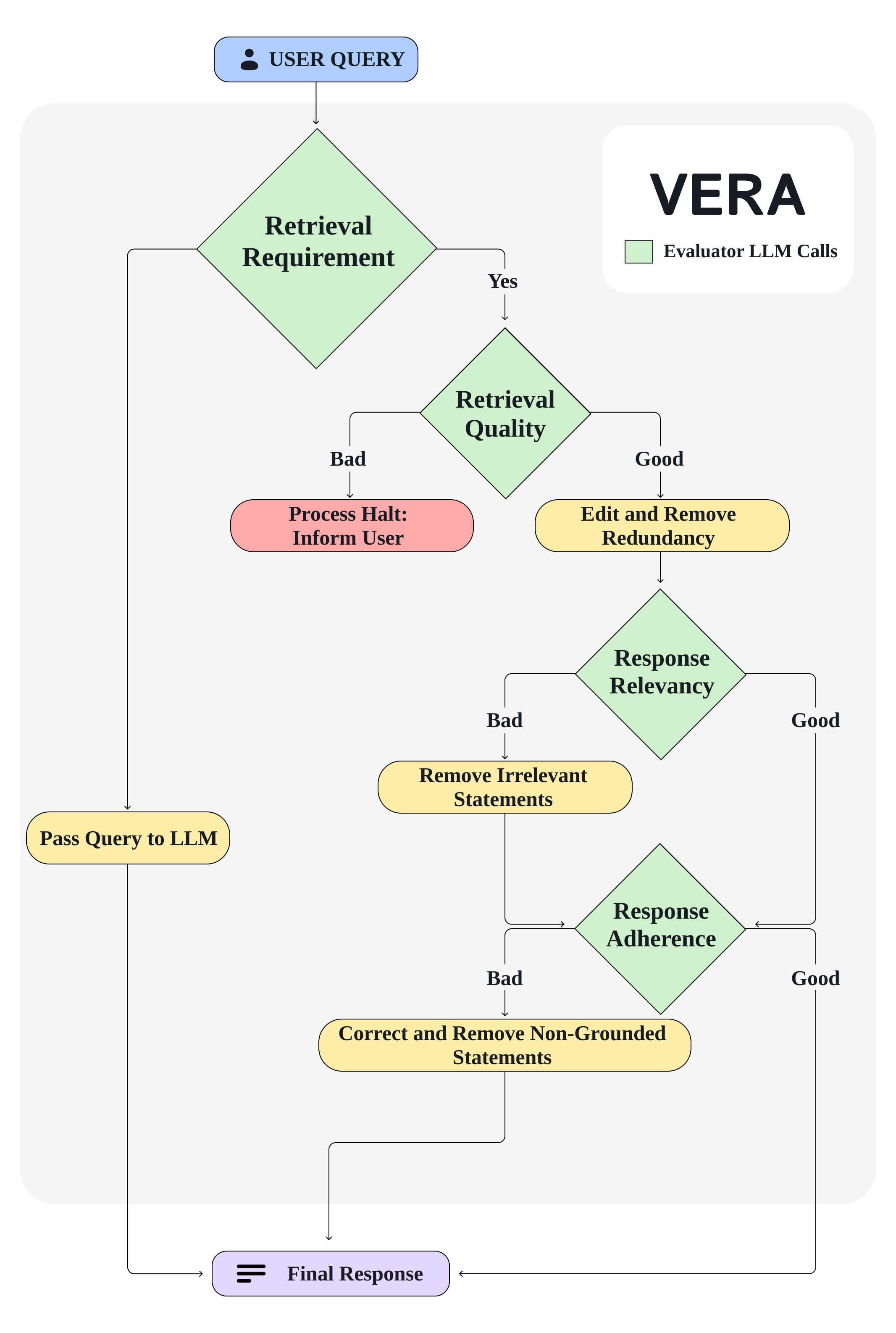}
    \caption{An overview of methodology of VERA}
\end{figure}
We present VERA, a fine-grained evaluator and enhancer for retrievers and LLMs within a RAG system. As depicted in the accompanying figure, VERA first evaluates and edits the retrieved context to increase its relevance and conciseness in relation to the query. This refined context is then provided to the LLM for response generation. After the response is generated, it undergoes further evaluation and editing to ensure it is concise and error-free, resulting in the final response.

All components of VERA are implemented using few-shot prompting. In all our experiments, we employ GPT-4o as the evaluator-cum-enhancer model due to its state-of-the-art capabilities.

\subsection{Retrieval Requirement check}

\begin{figure}[ht]
    \centering
    \fbox{\includegraphics[width=\columnwidth]{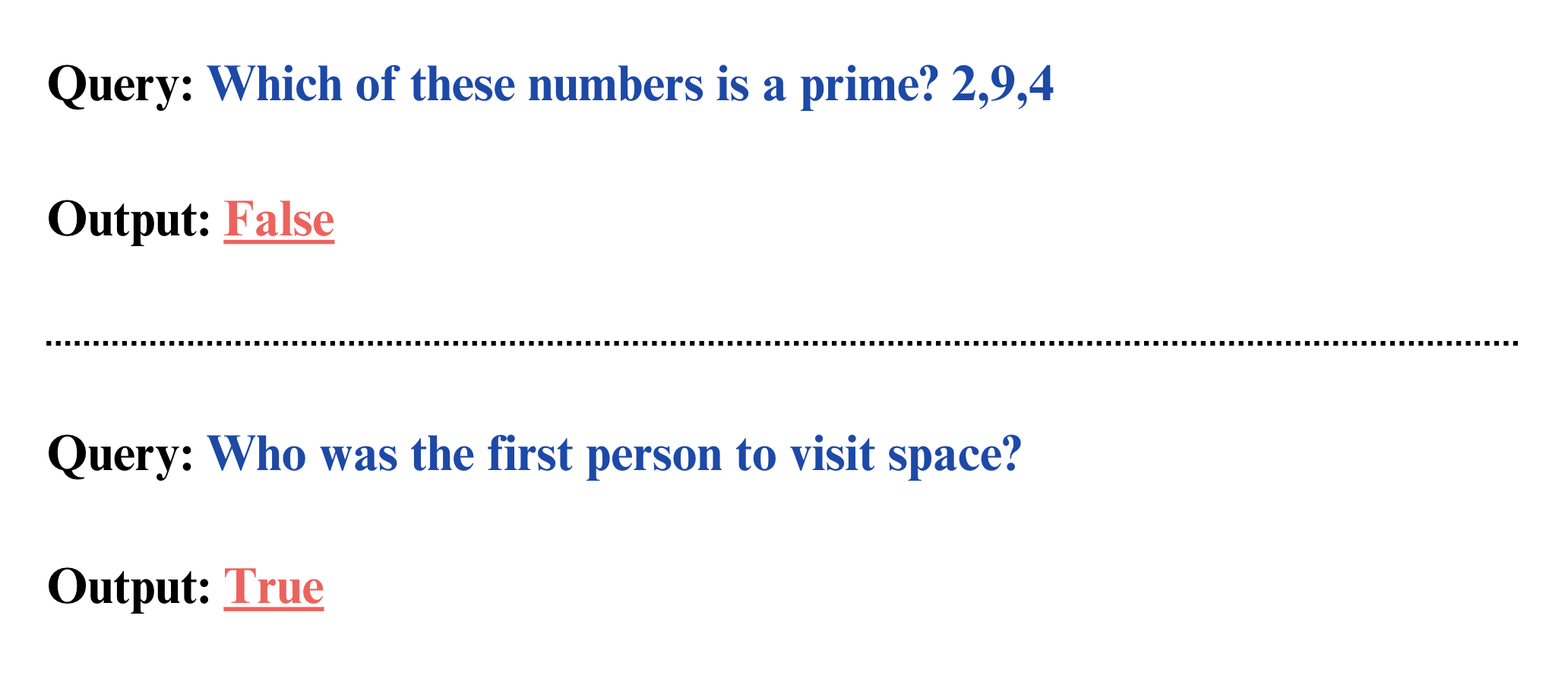}}
    \caption{Retrieval Requirement Check}
\end{figure}

\begin{algorithm}
\caption{Retrieval Requirement Check}
\textbf{Input:} User query $Q$ \\
\textbf{Output:} Boolean indicating if retrieval is needed 
\begin{algorithmic}[1]
\Function{NeedsRetrieval}{$Q$}
    \If{$Q$ \text{is knowledge-intensive}}
        \State \textbf{return} \text{True}
    \Else
        \State \textbf{ return} \text{False}
    \EndIf
\EndFunction

\end{algorithmic}
\end{algorithm}

Not all queries necessitate retrieval; only those that are knowledge-intensive do. Upon receiving a user prompt, VERA determines whether external context is required to answer the prompt or if it can be addressed using the model's internal knowledge. If retrieval is necessary, VERA proceeds to retrieve the required context. Otherwise, the prompt is passed directly to the LLM for response generation.

\subsection{Retrieval Quality Evaluation and Correction}
\begin{figure}[ht]
    \centering
    \fbox{
    \includegraphics[width=\columnwidth]{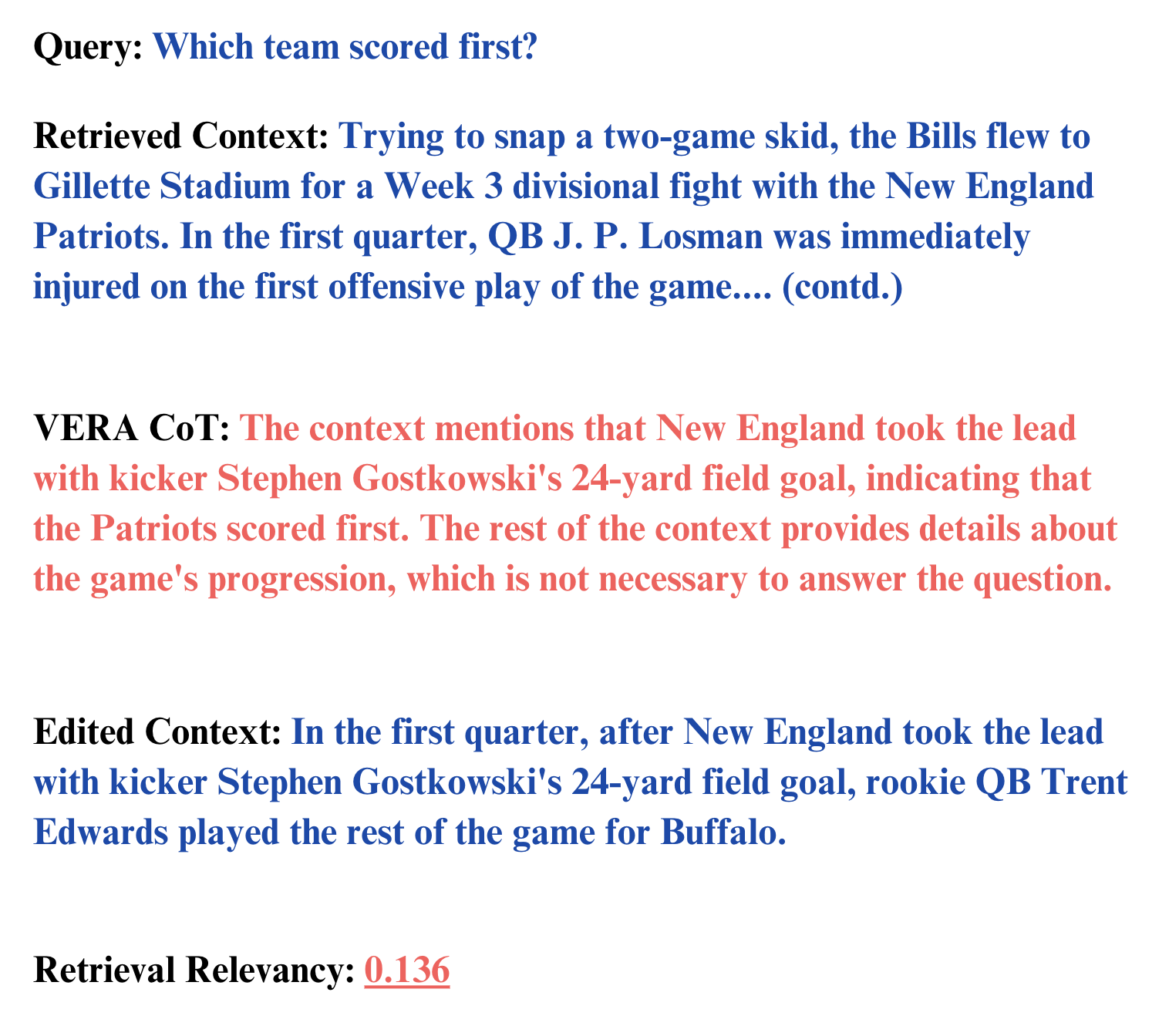}
}    \caption{Retrieval Quality Evaluation and Correction}
\end{figure}
\begin{algorithm}
\caption{Retrieval Quality Evaluation and Correction}
\textbf{Input:} Retrieved context $C$ \\
\textbf{Output:} Edited context $C'$
\begin{algorithmic}[1]

\Function{EvaluateAndEditContext}{$C$}
    \State $C' \gets \text{EliminateRedundantInformation}(C)$
    \State $R_{\text{retrieval}} \gets \frac{|C'|}{|C|}$
    \If{$|C'| = 0$} 
        \State \textbf{return} Query cannot be answered with retrieved context
    \Else
        \State \textbf{ return} $C'$
    \EndIf
\EndFunction

\end{algorithmic}
\end{algorithm}

After the retriever system retrieves the necessary context, VERA evaluates its relevance. VERA then edits the context to eliminate any redundant information that would not aid in answering the query without changing any other details or style. 

Let \( C \) be the original context retrieved by the retriever system and \( C' \) be the edited context after VERA has eliminated redundant information.

The retrieval relevance score \( R_{\text{retrieval}} \) is given by the ratio of the length of the edited context \( |C'| \) to the length of the original context \( |C| \):

\[
R_{\text{retrieval}} = \frac{|C'|}{|C|}
\]

If \( R_{\text{retrieval}} = 0 \) (i.e., \( |C'| = 0 \)), it indicates that the retrieved context fails to provide any useful information, and the process is halted. The user is then informed that their query cannot be answered. If \( R_{\text{retrieval}} > 0 \), it indicates that there is sufficient information in the context, and this edited context \( C' \) is used to generate the LLM's response to the user query.

\subsection{Response Relevancy Evaluation and Correction}
\begin{figure}[ht]
    \centering
    \fbox{
    \includegraphics[width=\columnwidth]{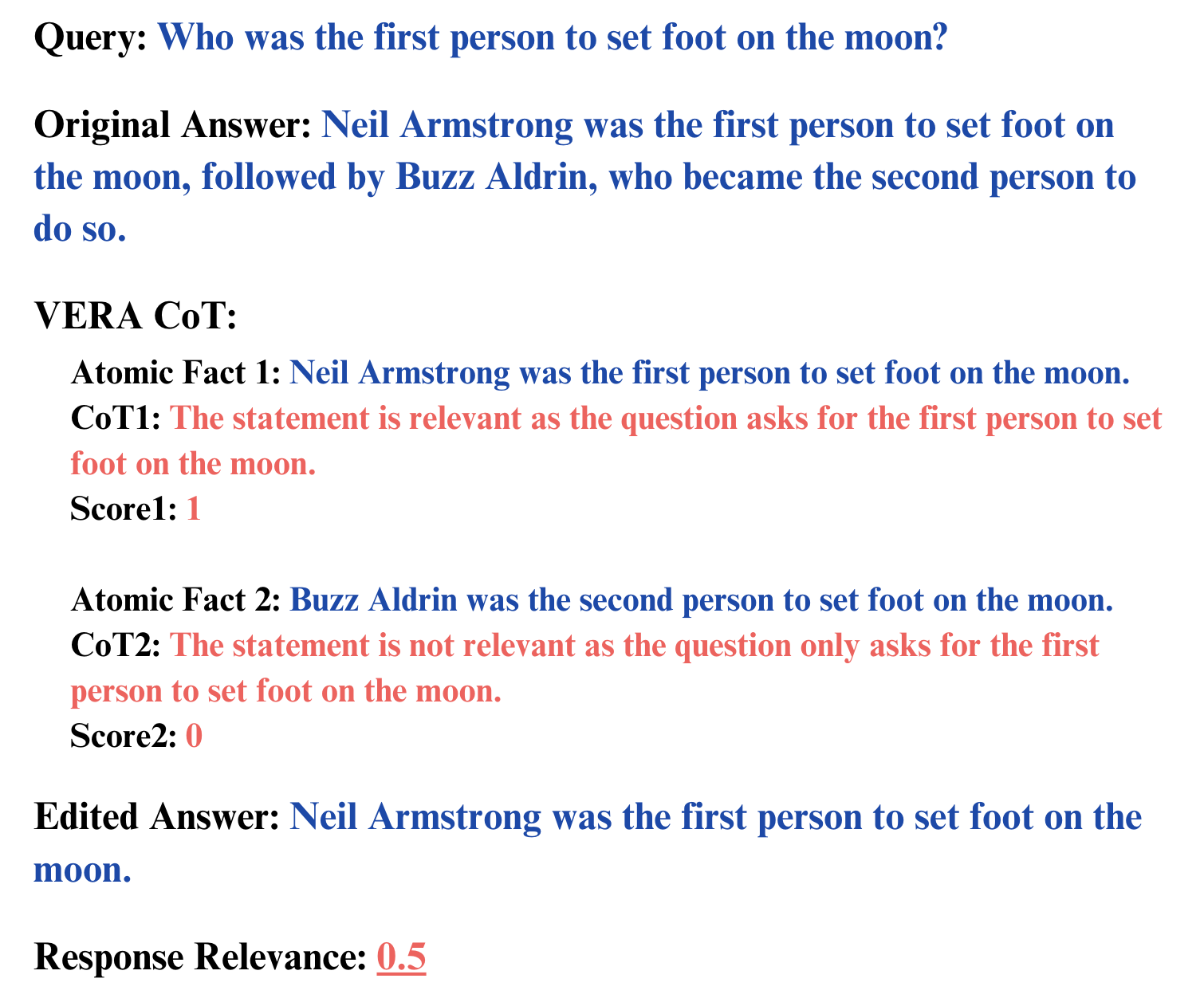}
}    \caption{Response Relevancy Evaluation and Correction}
\end{figure}
\begin{algorithm}
\caption{Response Relevancy Evaluation and Correction}
\textbf{Input:} Generated response $R$, User query $Q$ \\
\textbf{Output:} Edited response $R'$

\begin{algorithmic}[1]

\Function{EvaluateAndEditResponseRelevancy}{$R, Q$}
    \State $S \gets \text{SplitIntoAtomicStatements}(R)$
    \State $S' \gets \emptyset$
    \ForAll{$s_i \in S$}
        \If{$\text{IsRelevant}(s_i, Q)$}
            \State $S' \gets S' \cup \{s_i\}$
        \EndIf
    \EndFor
    \State $R_{\text{response}} \gets \frac{1}{|S|} \sum_{i=1}^{|S|} r(s_i)$
    \State \textbf{return} \text{JoinStatements}(S')
\EndFunction

\end{algorithmic}
\end{algorithm}

To ensure that the generated response contains only information pertinent to answering the query, VERA  evaluates and edits the response to eliminate any superfluous details. This process involves splitting the response into atomic statements and assessing the relevance of each statement in addressing the query using reasoning. Irrelevant atomic statements are removed from the original response while ensuring that the remaining content is preserved. This meticulous approach guarantees that the final response is concise and focused, devoid of any unnecessary information, thereby enhancing the overall quality and accuracy of the answer provided. 

Let \( S = \{s_1, s_2, \ldots, s_n\} \) be the set of atomic statements in the response. Each statement \( s_i \) is assigned a binary score \( r(s_i) \), where \( r(s_i) = 1 \) if the statement is relevant and \( r(s_i) = 0 \) if it is not.

The final response relevance score \( R_{\text{response}} \) is given by:

\[
R_{\text{response}} = \frac{1}{n} \sum_{i=1}^{n} r(s_i)
\]

where \( n \) is the total number of atomic statements in the response. This score reflects the proportion of the original response that is relevant to the query.

\subsection{Response Adherence Evaluation and Correction}
\begin{figure}[ht]
    \centering
    \fbox{
    \includegraphics[width=\columnwidth]{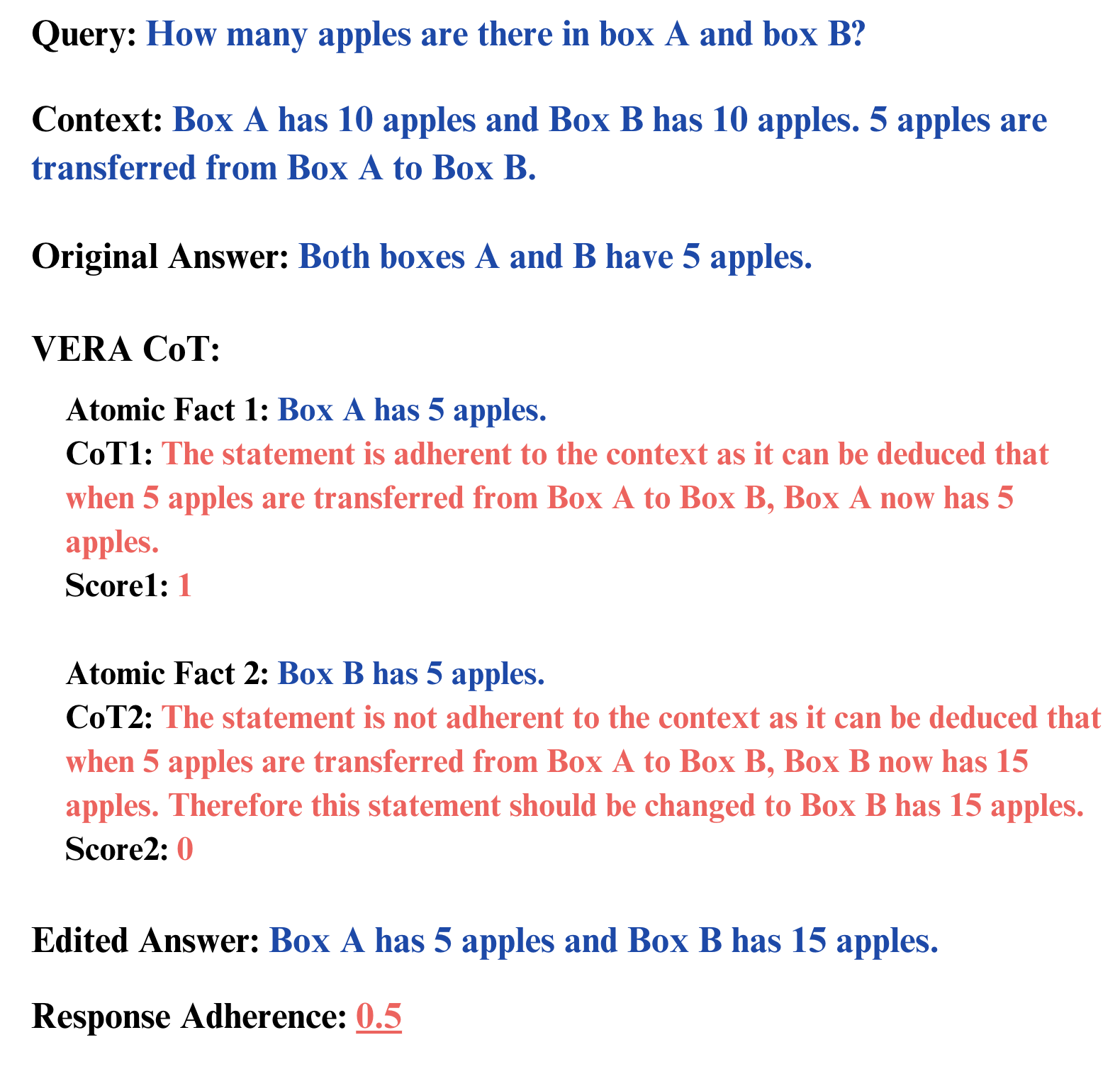}
}    \caption{Response Adherence Evaluation and Correction}
\end{figure}
\begin{algorithm}
\caption{Response Adherence Evaluation and Correction}
\textbf{Input:} Edited response $R$, Edited context $C'$ \\
\textbf{Output:} Final response $R'$
\begin{algorithmic}[1]
\Function{EvaluateAndEditResponseAdherence}{$R, C'$}
    \State $S \gets \text{SplitIntoAtomicStatements}(R)$
    \State $S' \gets \emptyset$
    \ForAll{$s_i \in S$}
        \If{$\text{IsGroundedInContext}(s_i, C')$}
            \State $S' \gets S' \cup \{s_i\}$
        \EndIf
    \EndFor
    \State $A_{\text{response}} \gets \frac{1}{|S|} \sum_{i=1}^{|S|} g(s_i)$
    \State \textbf{return} \text{JoinStatements}(S')
\EndFunction

\end{algorithmic}
\end{algorithm}

As discussed earlier, LLMs in RAG system may generate text that includes additional information beyond the retrieved documents \citep{shuster2021retrieval}, disregards the documents altogether, or even contradicts the documents. This was observed by us even in state-of-the-art LLMs like GPT-4o. VERA addresses this by splitting the response from the previous step (relevancy correction) into atomic statements similar to what is proposed in FactScore \citep{min2023factscore} and then assessing each of them. However, this approach of using a binary score to classify each statement as adherent or non-adherent yielded sub-optimal evaluation accuracy, as some statements, while not explicitly present in the context, could be logically inferred and should therefore be classified as adherent. To improve accuracy, we propose a more nuanced classification system for atomic statements, prompting the evaluator to categorize them into three distinct classes: (1) directly derivable from the context, (2) not directly derivable but logically inferable from the context, and (3) entirely inaccurate and not grounded in the context. This classification process is guided by chain-of-thought reasoning \citep{wei2022chain} to maximize precision. VERA then uses reasoning to make necessary edits by correcting any incorrect statements and removing statements which are not grounded in the context. 

The response adherence score is calculated by assigning a binary score to each atomic statement. If a statement is grounded in the context or deducible from the context, it is assigned a score of 1; otherwise, it receives a score of 0. The final response adherence score \( A_{\text{response}} \) is given by:

\[
A_{\text{response}} = \frac{1}{n} \sum_{i=1}^{n} g(s_i)
\]

where \( S = \{s_1, s_2, \ldots, s_n\} \) is the set of atomic statements in the response, \( g(s_i) \) is the binary score for each statement \( s_i \) (1 if grounded and accurate, 0 otherwise), and \( n \) is the total number of atomic statements in the response. This score reflects the proportion of the initial response that is accurate and adherent to the context.

\section{Experiments}

\subsection{Tasks and Datasets}
We rigorously assess VERA's effectiveness across various datasets and downstream tasks \citep{kucharavy2024adapting}. Our tests are designed to establish a fair baseline and accurately reflect real-world scenarios.

\subsubsection{SQuAD-2.0 Dataset} 
Stanford Question Answering Dataset (SQuAD) \citep{rajpurkar2016squad} is a reading comprehension dataset, consisting of questions posed by crowdworkers on a set of Wikipedia articles, where the answer to every question is a segment of text, or span, from the corresponding reading passage, or the question might be unanswerable. This dataset is challenging as there are questions that might not be answerable from the provided context.

\subsubsection{DROP Dataset}
The DROP dataset \citep{dua2019drop} serves as a reading comprehension benchmark designed for Discrete Reasoning Over Paragraphs. Comprising 96,000 questions, this dataset was adversarially crowd-sourced to challenge systems in a variety of tasks. To successfully navigate DROP, a system must interpret references within a question—potentially across multiple parts of the input—and carry out discrete operations such as addition, counting, or sorting. These tasks demand a thorough understanding of the paragraph's content.

\subsubsection{Real World Downstream Tasks}
To evaluate the effectiveness of VERA on real-world downstream tasks, we compiled a set of three documents representing diverse use cases of a RAG based LLM. These documents include:

\begin{enumerate}
\item \textbf{World War II Wikipedia Page:} The Wikipedia article on World War II presents a challenging evaluation, testing the model's capacity to adhere to the provided context without deviating due to its pre-existing knowledge from prior training.
\item \textbf{Apple 10-K Report:} The 2023 fiscal year Form 10-K for Apple was chosen to assess the RAG system's ability to handle numerical and financial data \cite{setty2024improving}, reflecting a common application of RAG models in processing and interpreting financial documents.
\end{enumerate}

\subsection{Baselines}
We assess publicly available pre-trained language models such as Mistral-7B-instruct-v0.1 \citep{jiang2023mistral}, GPT-3.5-turbo \citep{brown2020language}, and GPT-4o \citep{openai2024gpt4technicalreport} to demonstrate VERA's effectiveness across different model sizes. Mistral-7B-instruct-v0.1 represents a smaller model, while GPT-4o exemplifies a state-of-the-art model. Additionally, we compare these with the 7B Self-Rag \citep{asai2023self} \citep{touvron2023llama2openfoundation} model available on HuggingFace.

For downstream tasks, we utilize FAISS \citep{douze2024faiss} as a vector store and use similarity search retrieval, setting the chunk size to 512 tokens and chunk overlap to 25 tokens. To ensure consistency, GPT-4o is used as the evaluator model for VERA in all tests. The answers generated by VERA are further evaluated using GPT-4o to obtain post-enhancement scores. The questions to create a QA dataset from the given documents were created using the ragas library \citep{es2023ragas}. There was an equal proportion of questions testing reasoning abilities and questions that required multiple contexts to answer.

The SQuAD-2.0 and DROP datasets do not require a retriever system, as they provide the context directly within the dataset itself.
\section{Results}

\begin{table*}[ht]
\centering
\begin{tabular}{|l|l|l|}
\hline
\textbf{} & \textbf{SQuAD2.0}& \textbf{DROP}\\
\hline
\textbf{mistral-7B-instruct-v0.1}& 0.416& 0.432\\
\hline
\textbf{gpt-3.5-turbo} & 0.490& 0.696\\
\hline
\textbf{gpt-4o} & 0.582& 0.816\\
\hline
 \textbf{selfrag 7B}& 0.302&0.234\\\hline
 \textbf{mistral-7B-instruct-v0.1 + VERA} & 0.582&0.752\\\hline
 \textbf{gpt-3.5-turbo + VERA}& 0.640&0.764\\\hline
 \textbf{gpt-4o + VERA}& 0.690&0.854\\\hline
\end{tabular}
\caption{SQuAD2.0 and DROP Results}
\label{tab:comparison1}
\end{table*}
\begin{table*}[ht]
\centering
\begin{tabular}{|l|l|l|l|l|}
\hline
& & \textbf{mistral-7B-instruct-v0.1} & \textbf{gpt-3.5-turbo} & \textbf{gpt-4o} \\
\hline
\multirow{3}{*}{\textbf{Without VERA}} & \textbf{Response Adherence} & 0.740& 0.862& 0.906\\
& \textbf{Response Relevance} & 0.761& 0.917& 0.920\\
& \textbf{Context Relevance} & 0.311& 0.308& 0.309\\
\hline
\multirow{3}{*}{\textbf{With VERA}} & \textbf{Response Adherence} & 0.911& \textbf{0.970}& 0.964\\
& \textbf{Response Relevance} & 0.927& \textbf{0.982}& 0.944\\
& \textbf{Context Relevance} & 0.876& 0.883& 0.872\\
\hline
\end{tabular}
\caption{Comparison of models with and without VERA - WWII Wikipedia}
\label{tab:comparison3}
\end{table*}
\begin{table*}[ht]
\centering
\begin{tabular}{|l|l|l|l|l|}
\hline
& & \textbf{mistral-7B-instruct-v0.1} & \textbf{gpt-3.5-turbo} & \textbf{gpt-4o} \\
\hline
\multirow{3}{*}{\textbf{Without VERA}} & \textbf{Response Adherence} & 0.828& 0.900& 0.943\\
& \textbf{Response Relevance} & 0.716& 0.935& 0.943\\
& \textbf{Context Relevance} & 0.412& 0.427& 0.396\\
\hline
\multirow{3}{*}{\textbf{With VERA}} & \textbf{Response Adherence} & 0.896& 0.950& \textbf{0.971}\\
& \textbf{Response Relevance} & 0.945& \textbf{0.984}& 0.972\\
& \textbf{Context Relevance} & 0.895& 0.871& 0.881\\
\hline
\end{tabular}
\caption{Comparison of models with and without VERA - Apple 10k Report}
\label{tab:comparison2}
\end{table*}
\begin{figure}[ht]
    \centering
    \fbox{
    \includegraphics[width=\columnwidth]{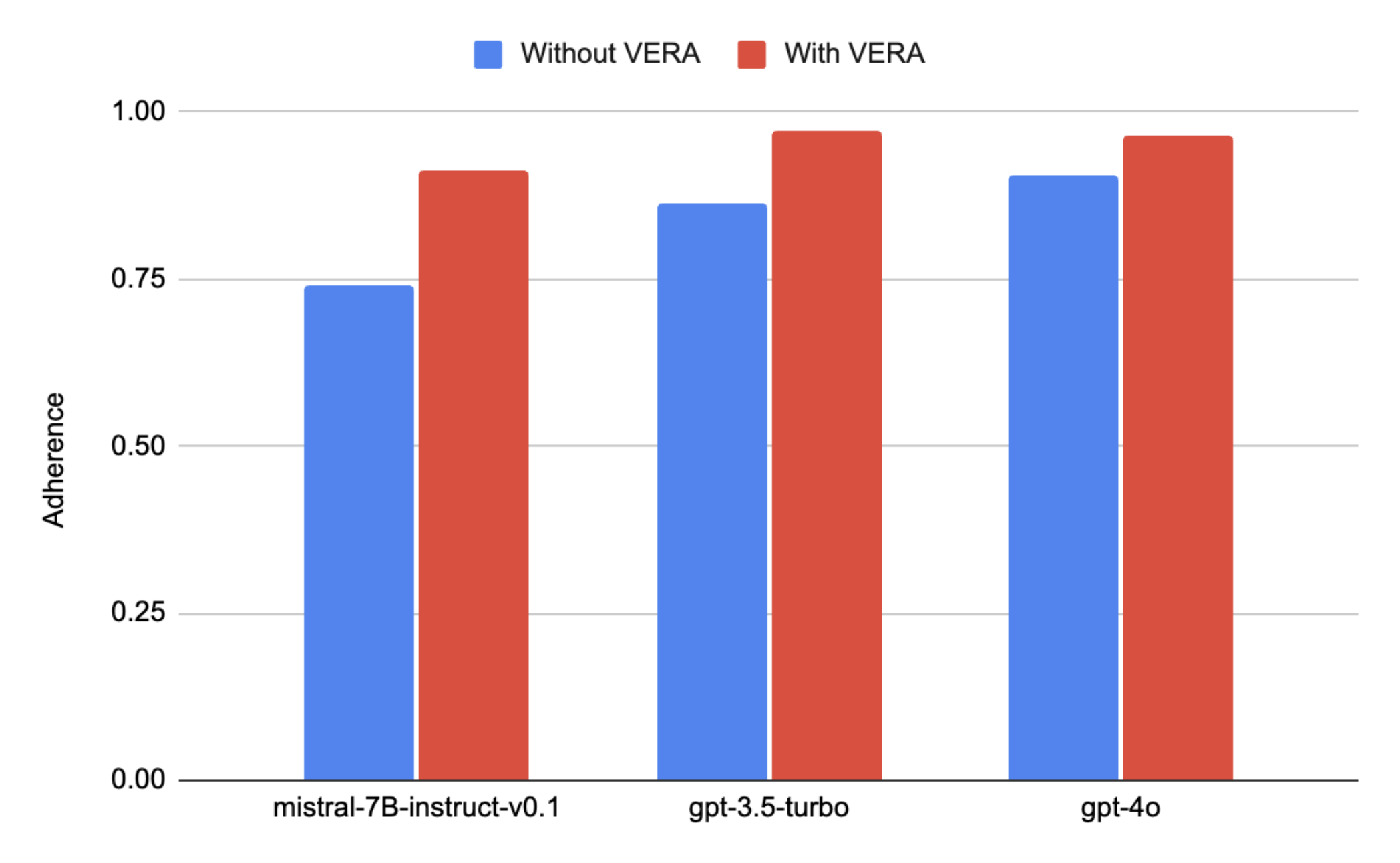}
  }  \caption{WWII Wikipedia Adherence Scores}
\end{figure}
\begin{figure}[ht]
    \centering
    \fbox{
    \includegraphics[width=\columnwidth]{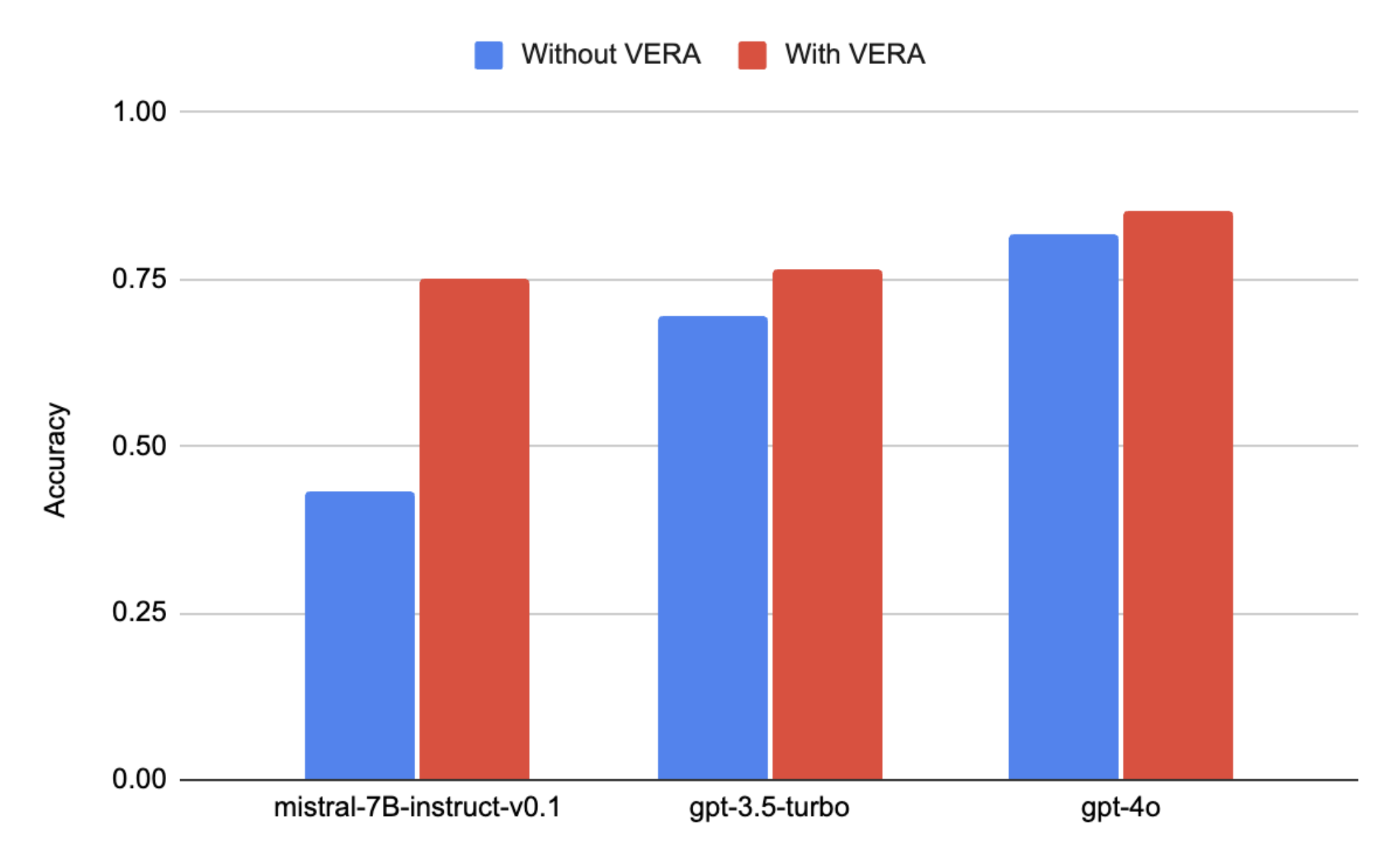}
}    \caption{DROP Accuracy}
\end{figure}
We observed a substantial improvement in accuracy for both the SQuAD2.0 and DROP datasets (Table 1) when employing VERA. Specifically, Mistral-7B-instruct-v0.1 exhibited a 20\% increase in accuracy on the SQuAD2.0 dataset and a 15\% increase on the DROP dataset. Additionally, VERA enhanced the performance of GPT-4o by 5\% on SQuAD2.0 and 10\% on DROP. These results underscore VERA's effectiveness in enhancing the performance of large language models on tasks that demand advanced comprehension capabilities.

The results of downstream tasks demonstrated a significant increase in adherence and relevance scores for smaller models like Mistral-7B-instruct-v0.1. Notable improvements were also observed in larger models such as GPT-4o and GPT-3.5-turbo. Specifically, Mistral-7B-instruct-v0.1 exhibited an increase in Response Adherence by up to 18.7\% (Table 2) and an increase in Response Relevance by up to 17.9\% (Table 2) when using VERA.

The improvements in Response Adherence and Relevance for GPT-4o indicate that VERA can be effectively used for self-improvement \citep{huang2022large}, as the evaluator model employed was also GPT-4o. This finding is significant because it demonstrates that VERA's performance enhancements are not solely attributable to the use of GPT-4o but rather to the systematic evaluation and refinement processes implemented by VERA.

In all the downstream tasks, the initial Context Relevance was below 0.45. This can be attributed to the larger chunk size of 512 tokens \citep{eibich2024aragog}, of which only approximately 30\% to 45\% of the information was relevant to the context. While Context Relevance is not directly dependent on the LLM used, we still observed variations in the scores due to the stochastic nature of the LLM serving as the evaluator \citep{sun2024comprehensive}. Despite this inherent variability, the use of VERA led to a clear and consistent increase in Context Relevance across all experiments.

\section{Conclusion}
In this work, we presented VERA, a novel system designed to address the limitations of Retrieval-Augmented Generation (RAG) in enhancing Large Language Models (LLMs). By incorporating an evaluator-cum-enhancer LLM, VERA significantly improves the relevance, adherence, and overall quality of responses. Our approach involves a multi-step process that determines the necessity of retrieval, evaluates and refines retrieved documents, and rigorously assesses and corrects the generated responses

VERA's method of breaking down responses into atomic facts and ensuring each statement's grounding in the retrieved context leads to higher fidelity and relevance in the final outputs. Our experimental results demonstrate that VERA increases adherence and relevance significantly for both smaller LLMs like Mistral 7B instruct v0.1 and larger models like GPT-4o, showcasing its versatility and effectiveness across different model scales.

The improvements brought by VERA highlight its potential in applications where accurate and reliable information generation is crucial. By mitigating hallucinations and refining the retrieval and response process, VERA paves the way for more trustworthy and contextually appropriate LLM outputs, advancing the state-of-the-art in retrieval-augmented language modeling.

\section{Limitations and Future Work}
VERA demonstrates strong capabilities in understanding semantic changes between the response and context, avoiding unnecessary penalties for semantically equivalent statements (e.g., "World War II is a deeply engraved event in history" and "World War II is an important event in history"). However, during our experimentation, we observed that smaller models like Mistral-7B-instruct or Llama3 8B, when used as evaluators instead of GPT-4o, struggled to handle such semantic nuances effectively. This limitation could potentially be addressed by improving the few-shot prompting technique, thereby enhancing evaluation performance with smaller models and making the method more cost-efficient.

Due to the stochastic nature of the evaluator LLM, the splitting of the response into atomic statements may vary slightly with each evaluation, resulting in minor differences in scores. Although this limitation is largely mitigated by using a large dataset in our experiments, it can still cause minor variations in scores for individual evaluations.

Since VERA necessitates LLM evaluation at each step of the process, it might not be suitable for real-time applications. This limitation could potentially be addressed in the future by combining multiple evaluation calls into a single step, thereby making the process more streamlined and time-efficient.

\bibliography{aaai25}

\end{document}